\documentclass[preprint,12pt]{elsarticle}

\usepackage{xcolor}
\usepackage{soul}

\usepackage{amssymb}
\usepackage{amsmath}
\usepackage{multirow}
\usepackage{booktabs}
\usepackage{longtable}
\usepackage{caption}
\usepackage{subcaption}
\usepackage{url}
\usepackage{lineno}
\usepackage{algorithm}
\usepackage{algorithmic}
\floatname{algorithm}{Pseudo-code}

\begin{document}

\begin{frontmatter}

\title{Explainable Admission-Level Predictive Modeling for Prolonged Hospital Stay in Elderly Populations
: Challenges in Low- and Middle-Income Countries}


\author[a,b]{Daniel Sierra-Botero}
\author[a,b]{Ana Molina-Taborda}
\author[c,d]{Leonardo Espinosa-Leal}
\author[a]{Alexander Karpenko}
\author[e,f]{Alejandro Hernandez}
\author[a,b]{Olga Lopez-Acevedo\corref{cor1}}

\cortext[cor1]{Corresponding author}

\ead{olga.lopeza@udea.edu.co}


\affiliation[a]{
  organization={Biophysics of Tropical Diseases Max Planck Tandem Group, University of Antioquia (UdeA)},
  addressline={Calle 67 \# 63-108},
  city={Medellín},
  postcode={050010},
  state={Antioquia},
  country={Colombia}
}

\affiliation[b]{
  organization={Grupo de Física Atómica y Molecular, Instituto de Física, University of Antioquia (UdeA)},
  addressline={Calle 67 \# 63-108},
  city={Medellín},
  postcode={050010},
  state={Antioquia},
  country={Colombia}
}

\affiliation[c]{
  organization={Graduate School and Research, Arcada University of Applied Sciences},
  addressline={Jan-Magnus Janssonin aukio 1},
  city={Helsinki},
  postcode={00560},
  country={Finland}
}

\affiliation[d]{
  organization={VTT Technical Research Centre of Finland Ltd},
  city={Espoo},
  postcode={02044-VTT},
  country={Finland}
}

\affiliation[e]{
  organization={Department of Internal Medicine, School of Medicine, University of Antioquia},
  city={Medellín},
  country={Colombia}
}

\affiliation[f]{
  organization={Hospital Alma Mater de Antioquia, University of Antioquia},
  city={Medellín},
  country={Colombia}
}

\begin{abstract}

Prolonged length of stay (pLoS) is a significant factor associated with the risk of adverse in-hospital events. We develop and explain a predictive model  for pLos using admission-level patient and hospital administrative data.
The approach includes a feature selection method by selecting non-correlated features with the highest information value. The method uses features weights of evidence to select a representative within cliques from graph theory. The prognosis study analyzed the records from 120,354 hospital admissions at the Hospital Alma Mater de Antioquia between January 2017 and March 2022. After a cleaning process the dataset was split into training (67\%), test (22\%), and validation (11\%) cohorts. A logistic regression model was trained to predict the pLoS in two classes: less than or greater than 7 days.
The performance of the model was evaluated using accuracy, precision, sensitivity, specificity, and AUC-ROC metrics. The feature selection method returns nine interpretable variables, enhancing the models' transparency.
In the validation cohort, the pLoS model achieved a specificity of 0.83 (95\% CI, 0.82-0.84), sensitivity of 0.64 (95\% CI, 0.62-0.65), accuracy of 0.76 (95\% CI, 0.76-0.77), precision of 0.67 (95\% CI, 0.66-0.69), and AUC-ROC of 0.82 (95\% CI, 0.81-0.83).
The model exhibits strong predictive performance and offers insights into the factors that influence prolonged hospital stays. This makes it a valuable tool for hospital management and for developing future intervention studies aimed at reducing pLoS.

\end{abstract}


\end{frontmatter}

\section{Introduction}

The functioning of a hospital relies on its administrative guidelines, operational tasks, and the severity of the patient's condition. A critical factor in resource utilization and the risk of adverse events is the length of stay (LOS), which refers to the number of days a patient remains in the hospital from admission until discharge \cite{Awad2017}. In low- and middle-income countries, social and economic inequalities lead to healthcare systems that often face limited hospital resources and staff shortages. Under these conditions, hospitals must assume the dual role of both acute care and long-term care providers for individuals who lack adequate social support or access to specialized care \cite{Orooji2021}. In particular, geriatric patients face unique challenges due to the scarcity of elder care facilities, which are often prevalent in high-income countries. In addition, when a patient's length of stay exceeds seven days, prolonged length of stay (pLOS), the risk of in-hospital complications increases 10-fold, affecting hospital costs. Although studies have implemented interventions to reduce LOS, their success depends on the specific conditions of each hospital \cite{Tipton2001, Leite2022}. For this reason, an accurate prediction of LOS and pLOS provides valuable information that may help improve the hospital's efficiency and reduce the risk of adverse events.

In some high-income countries, machine learning models have been developed to predict LOS \cite{Zang2023-bz}. Some of them were focused on Intensive Care Unit (ICU) facilities \cite{Wu2021}. In contrast, others had a particular interest in studying LOS in cardiac affections \cite{Rotar2022, Meadows2018} or traumatic injuries and fractures \cite{Choi2023, Liu2024}. Increasing the population to all hospitalized patients with various diseases remains a challenge \cite{Zang2023-bz}. Despite the good performance of these models, they do not adequately account for the unique challenges faced by resource-limited general hospitals that serve low- and middle-income elderly populations. Thus, predictive models tailored to these unique social regions are necessary to support general hospitals in managing. Additionally, having interpretable models allows one to understand the inner mechanics of the hospital, rather than relying on a black box prediction; that’s why some studies include explainability analysis of input variables through Shapley values\cite{Choi2023}.

This study aims to develop and validate an interpretable prediction model for pLoS using admission-level data from a hospital in a low- and middle-income country. It is focused on the elderly population, typically managed by general hospitals, who are admitted for any illness or condition. The model is designed for use by hospital administrators and healthcare professionals to support critical decisions regarding resource allocation, bed management, potential risk mitigation interventions, and diseases. The feature selection method is based on the weight of evidence and cliques from graph theory. This enhances the interpretability of factors related to hospital patient length of stay, facilitating informed decision-making for hospital administrators. It is validated by performance metrics such as accuracy, precision, recall, and the area under the receiver operating characteristic curve (AUC-ROC). The model’s interpretability is enhanced by a feature selection method and its comparison with Shapley values.

\section{Theory and Methods}

\subsection{Data Source}

The population in this study comes from hospital records of admitted patients at the Hospital Alma Mater de Antioquia, covering the period from January 2017 to March 2022. The Hospital Alma Mater is a mixed public-private corporation owned by the Universidad de Antioquia and the Fundación de Apoyo a la Universidad de Antioquia. The data involved 120,354 records used for administrative purposes. It included numerical and categorical features, allowing the prediction model to be trained on a diverse and large population. The dataset contained records with various types of admission level variables like ages, diseases, insurance types, and other characteristics, thereby increasing the model’s generalization and applicability for any patient admitted to the hospital.

Missing outcomes were processed in different ways: for numerical variables, a criterion for exclusion was applied (16,741 admissions), while for categorical variables with missing outcomes, an internal category labeled "missing outcome" was added, and these were retained in the dataset. No age restrictions were applied to the eligible patients; however, patients younger than 1 year or older than 90 years were assigned the ages of 0 years and 90 years, respectively. There were no treatments received by the patients before, at the beginning of, or during the study period. All the data available from the hospital from the end of 2017 until the beginning of 2023 were used in the study. 

Starting from the discharge database, the features were reduced to those collected during the admission process. Afterward, a feature selection process was performed using the Weight of Evidence (WoE) and Information Value (IV) methods, resulting in the final set of features used in the model.

\begin{figure}
    \centering
    \includegraphics[width=\textwidth]{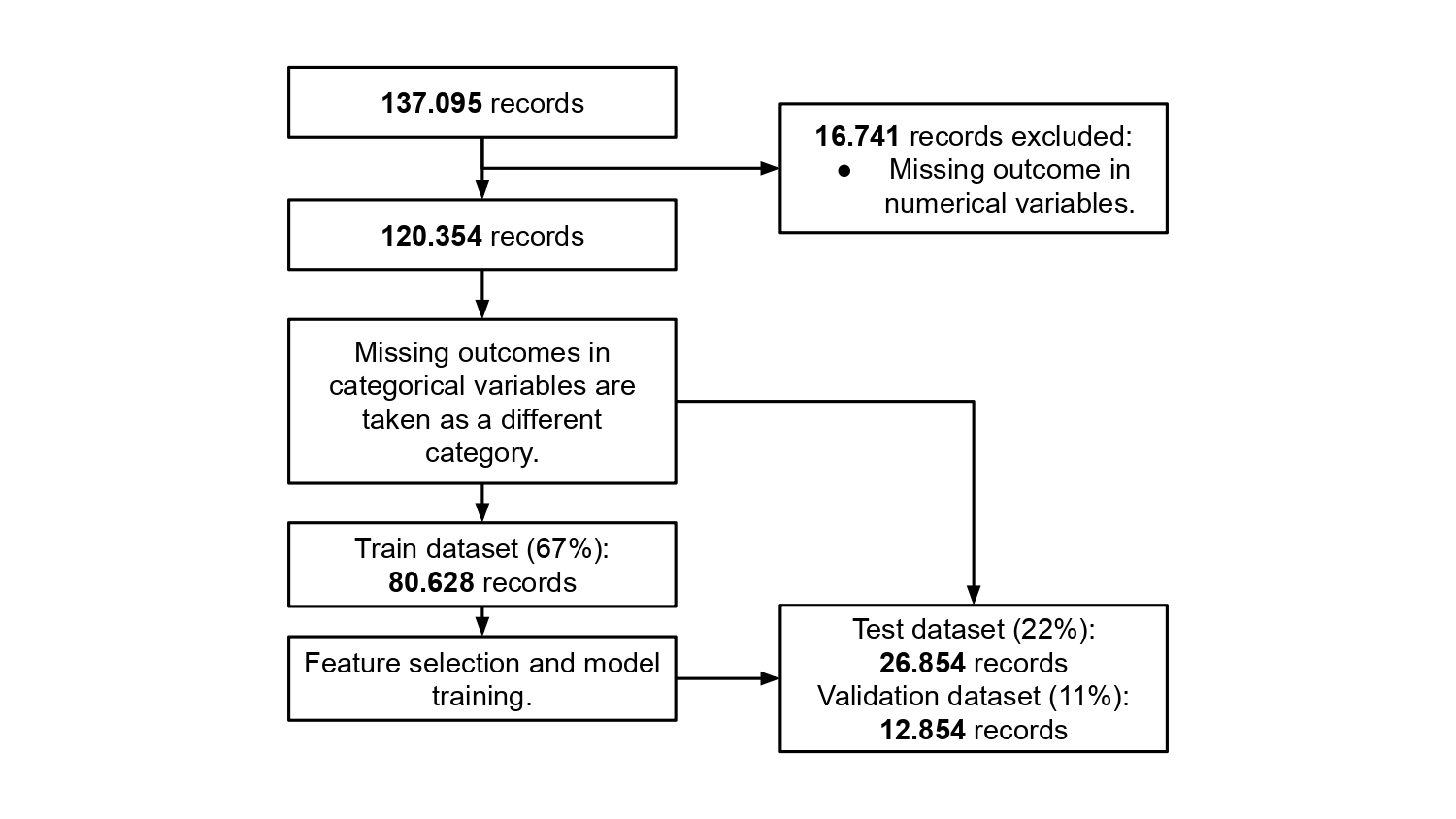}
    \caption{Workflow of data preprocessing, feature selection, and model development. The figure shows the number of admissions at each stage of the pipeline, starting from the original hospital discharge database and after data cleaning and missing outcome handling. The dataset was restricted to features available at admission and reduced through a feature selection process based on Weight of Evidence (WoE) and Information Value (IV). The final cohort was split into training (67\%), test (22\%), and validation (11\%) sets, and used for model training and evaluation.}
    \label{fig:workflow}
\end{figure}

The data split was as follows: the training dataset (67\%) and test dataset (22\%) correspond to the period from December 2017 to mid-September 2022, while the validation dataset (11\%) corresponds to the period from mid-September 2022 to March 2023. Admissions related to the same patient (8,156 patients) in the training dataset were processed by randomly selecting one admission and placing the rest in the test set. Figure \ref{fig:workflow} summarizes the complete flow of the patient data.

We used Python 3.9 (Python Software Foundation) and the Pandas library version 2.2.1 to remove missing outcomes, add the "missing outcome" category, and split the dataset. The models were trained using Scikit-Learn version 1.4.2.

\subsection{Proposed feature selection method}

A Python class was created to perform feature selection before model training. The class first calculates the Weight of Evidence equation \ref{eq:woe} and Information Value equation \ref{eq:iv} for each variable. The WoE is a measure that encodes categorical variables based on the distribution of the binary target variable. The WoE for a given category $i$ is given by:

\begin{equation}
\label{eq:woe}
\text{WoE}_i = \ln\left(
\frac{\text{Positive}_i / \text{Total Positives}}
{\text{Negative}_i / \text{Total Negatives}}
\right)
\end{equation}

where $Positive_{i}$ and $Negative_{i}$ represent the counts of the positive and negative target outcomes in category $i$, and $Total Positive$ and $Total Negative$ are the total counts across all categories. In other words, the WOE is the logarithm of the odds ratio of the positive outcomes to negative outcomes.

The IV quantifies the predictive power of a variable with several categories. It helps to identify which variables are most predictive of the binary outcome. The IV for a variable is calculated as:

\begin{equation}
\label{eq:iv}
\text{IV} = \sum_{i} \left(
\frac{\text{Positive}_i}{\text{Total Positive}} -
\frac{\text{Negative}_i}{\text{Total Negative}}
\right) \times \text{WoE}_i
\end{equation}

Where the sum of the products of the WoE and the difference in proportions of positive and negative outcomes for each category in the variable \cite{Good1975}.

The categorical variables were all transformed into numerical variables by assigning them their WOE value. Once the transformation was performed and all variables had been assigned discrete numerical values, a preliminary check was made to ensure that no variables had variance lower than $0.03$. 

A correlation matrix depicting the correlation between all the possible pairs of variables is obtained. An undirected graph is then constructed to represent the internal correlations among the variables. Vertices represent each variable, and edges are generated between pairs of variables with a correlation greater than 0.5. Once the graph is constructed, all maximal cliques (subsets of vertices, all adjacent to each other, to which no more vertices can be added) are found using the library `networkx` \cite{hagberg2008exploring}. As an output of this step, we obtain subsets of highly correlated variables. 

Since the main objective is to build an interpretable and mathematically simple model, it is essential to feed the model with the most relevant variables a strategy that in general prevents overfitting  \cite{Murphy2012}. Therefore, for feature selection, a representative variable from each clique is chosen using IV as a score, selecting the variable with the highest IV. If a representative belongs to more than one clique, it is discarded if a better representative exists. Also, representative variables with an IV less than 0.1 are removed, discarding variables that provide little to no predictive information. 

Finally, a transformation was performed on the set of representative variables, using the `OptimalBinning` class from the `OptBinning` library \cite{optbinning}, which groups different values of each variable into subgroups or ranges to obtain the IV and then identifies the arrangement that maximizes the IV parameter with respect to the target variable. This generates a new discrete variable where each optimized subgroup of categories is assigned its respective WoE. 

This optimized set of representative variables is used to train the prediction models.

We summarize the proposed feature selection method with the following pseudo-code:
\begin{algorithm}
\caption{Feature Selection by Information Value and Clique Reduction}
\label{feature_selection}
\begin{algorithmic}
\REQUIRE Input feature set $X = \{x_1, x_2, \dots, x_n\}$, target variable $y$
\ENSURE Reduced set of features

\STATE Remove low variance variables

\STATE Initialize empty list $IV\_scores$
\FOR{each feature $x_i \in X$}
    \IF{$x_i$ is numerical}
        \STATE Compute Information Value $IV(x_i,y)$
    \ELSIF{$x_i$ is categorical}
        \STATE Apply optimal binning on $x_i$
        \STATE Compute Information Value $IV(x_i,y)$
    \ENDIF
    \STATE Store $IV(x_i,y)$ in $IV\_scores$
\ENDFOR

\STATE Partition features into cliques $\mathcal{C} = \{C_1, C_2, \dots, C_m\}$

\STATE Initialize empty set $W$ (winners)
\FOR{each clique $C_j \in \mathcal{C}$}
    \STATE Find winner feature $x^* = \arg\max_{x \in C_j} IV(x,y)$
    \STATE Add $x^*$ to $W$
    \STATE Remove all other features in $C_j$ from consideration
\ENDFOR

\RETURN $W$ (set of winning features)
\end{algorithmic}
\end{algorithm}

\subsection{Model Development}

The model trained was a multivariate logistic regression with features selected by the method presented in the previous section. We tuned the hyperparameters using a grid search along the L2 ratio penalty and the inverse of regularization $C$. Each record used in the training dataset belongs to a different patient.

\subsection{Statistical Analysis}

Descriptive statistics were calculated for the study population to summarize key characteristics. Continuous variables, such as age, were presented as means with standard deviations (SD), while categorical variables, like sex and hospitalization type, were reported as counts and percentages. 

The significance of differences between subgroups was assessed using p-values, with a threshold of $p<0.05$ considered statistically significant. These descriptive methods were used to characterize the population and to detect any potential biases or imbalances between the groups, as shown in Table \ref{tab:characteristics}.

In addition to descriptive statistics, model performance was evaluated using accuracy, precision, recall, F1 score, and the area under the receiver operating characteristic curve (AUC-ROC). Bootstrap resampling was employed to calculate 95\% confidence intervals (CIs) for each performance metric.

The calibration of the model was assessed using a calibration curve, and the slope was analyzed to determine the degree of alignment between the predicted probabilities and the actual outcomes. Statistical significance was evaluated using p-values for the descriptive comparisons and other analysis steps.

\begin{table}
\scriptsize
    \centering
    \begin{tabular}{|p{2.2cm}|p{4.0cm}lp{1.6cm}p{1.6cm}l|} 
    \hline
\textbf{Category} & \textbf{Subcategory}  & \textbf{Overall}  & \textbf{Length of Stay \ensuremath{<} 7d}   & \textbf{Length of Stay \ensuremath{>} 7d}            & \textbf{P-Value} \\ [0.5ex]
\hline\hline

\textbf{Sex} & & & & & \ensuremath{<}0.001 \\
& Female   & 40289 (50.0) & 28574 (50.8) & 11715 (48.1) &  \\
   & Male            & 40339 (50.0) & 27684 (49.2) & 12655 (51.9) &           \\
   \hline
\textbf{Age} & & & & & \ensuremath{<}0.001 \\
&Mean   & 55.8  & 53.1   & 62.2  &   \\
& St. dev.  & 24.5  & 25.7  & 20.3  &     \\
\hline
\textbf{Hospitalization Type} & & & & &  \ensuremath{<}0.001 \\

& General adults                                    & 57125 (71.1) & 36561 (65.3) & 20564 (84.4) &  \\
& Emergency room \ensuremath{>}= 24 hours
&  12735  (15.9)  & 12078 (21.6) & 657 (2.7) &  \\
& General pediatric                              
& 4070 (5.1)   & 3625 (6.5)   & 445 (1.8)    &  \\
& Intensive care adults                             &  3896 (4.9)   & 1699 (3.0)   & 2197 (9.0)  &  \\
& Intermediate care adults                        
& 1557 (1.9)   & 1058 (1.9)   & 499 (2.0)    &  \\ 
& Emergency room \ensuremath{<} 24 hours
&  933  (1.2)    & 933  (1.7) &              &  \\
& Missing & 312 & & & \\  
\hline
\textbf{Prevalent Diagnostics} & & & & &  \ensuremath{<}0.001 \\

& Injury, poisoning, and certain other causes 
& 12959 (17.1) & 10264 (19.1) & 2695 (12.4)  &\\
& Diseases of the respiratory system            
& 11566 (15.3) & 7638 (14.2)  & 3928 (18.0)  &\\
& Diseases of the skin and subcutaneous tissue 
& 11200 (14.8) & 8698 (16.2)  & 2502 (11.5)  &\\
& Neoplasms                                       
& 8481 (11.2)  & 4146 (7.7)   & 4335 (19.9)  &\\
& Diseases of the digestive system         
& 6852 (9.1)   & 4824 (9.0)   & 2028 (9.3)   &\\
& Pregnancy, childbirth and the puerperium          & 7401 (9.8)   & 6304 (11.7)  & 1097 (5.0)   &\\
& Diseases of the nervous system
& 3899 (5.2) & 3178 (5.9) & 721 (3.3)        &\\
& Certain infectious and parasitic diseases
& 3371 (4.5) & 2178 (4.0) & 1193 (5.5)       &\\
& Diseases of the genitourinary system
& 3123 (4.1) & 2032 (3.8) & 1091 (5.0)       &\\
& Endocrine, nutritional, and metabolic diseases
& 2971 (3.9) & 1564 (2.9) & 1407 (6.5)       &\\
& Mental and behavioral disorders
& 1498 (2.0) & 1324 (2.5) & 174 (0.8)        &\\
& Diseases of the musculoskeletal system
& 1376 (1.8) & 880 (1.6) & 496 (2.3)         &\\
& Diseases of the circulatory system
& 341 (0.5) & 326 (0.6) & 15 (0.1)           &\\
& Diseases of the eye and adnexa
& 324 (0.4) & 251 (0.5) & 73 (0.3)           &\\
& Symptoms, signs, abnormal findings
& 140 (0.2) & 96 (0.2) & 44 (0.2)            &\\
& Certain conditions originating in the perinatal period
& 84 (0.1) & 78 (0.1) & 6 (0.0)              &\\
& Congenital malformations, deformations, and chromosomal abnormalities
& 24 (0.0) & 23 (0.0) & 1 (0.0)              &\\

&Missing&5018&&&\\
\hline
Total  &              & 80628        & 56258       & 24370       &           \\
\hline
\end{tabular}
\caption{Characteristics of the studied population. Percentages indicated in parenthesis in each subcategory is with respect to the total population indicated in the final row discriminated in overall, short and long stays.}
\label{tab:characteristics}
\end{table}

\subsection{Ethical approval and others}
According to local ethical standards (Resolution 8430 of 1993, Colombian Ministry of Health and Social Protection), the research is classified as having no risk. This includes data analysis derived from medical records or images, which does not involve direct patient contact. 

This research study does not involve prospective enrollment of subjects. According to local legislation, the consent can be waived for the use of retrospective data.

All data was stored in electronic format and housed in a secure server.  Data was extracted and de-identified. All scientific products resulting from this research do not identify any subjects and present data in aggregate form only.

\section{Results}

\subsection{Statistics of the population}

The study population consisted of 80,628 patients, with an equal distribution of sexes: 50.0\% male (40,289) and 50.0\% female (40,339). The mean age of the population was 55.8 years (SD: 24.5), with a significant difference between subgroups $(p < 0.001)$. The breakdown of hospitalization types revealed that the majority of patients were general adults (71.1\%), with smaller percentages for emergency room stays longer than 24 hours (15.9\%), emergency room stays shorter than 24 hours (1.2\%), pediatric patients (5.1\%), intensive care unit (ICU) adults (4.9\%), and intermediate care adults (1.9\%).

 In terms of primary diagnoses, the most prevalent categories were diseases of the respiratory system, which accounted for 15.3\% of the population (11,566 patients). Injury, poisoning, and other consequences of external causes represented 17.1\% (12,959 patients). Diseases of the skin and subcutaneous tissue made up 14.8\% (11,200 patients), while neoplasms were present in 11.2\% of cases (8,481 patients). Lastly, diseases of the digestive system comprised 9.1\% (6,852 patients) of the diagnoses.

 Patients were predominantly treated for conditions related to the respiratory system, injuries, skin diseases, and cancer, indicating a broad range of medical conditions present in this population. Significant differences in these diagnostic distributions were noted across age and sex subgroups $(p < 0.001)$. Additionally, hospital stays were distributed across different service areas, with patients admitted to special care units (e.g., ICU) comprising 9.0\% of the population, highlighting the relevance of critical care within this dataset.

\subsection{Data, Input variables and Model}

As described in the Methods section, all feature variables were preprocessed to ensure they contained sufficient information and low correlation levels between them. The admission-level initial variables included 79 features.
Inserted into the search cliques algorithm results on 41 cliques or sets of variables that are highly correlated.  To avoid multicollinearity, only the representative variable from each clique with the highest Information Value (IV) score was retained, resulting in a final set of 9 variables.

The selected input variables for the model included the main diagnosis (main dx), related diagnosis (r3 dx), presurgical diagnosis (presr dx), antibiotic use (antibiotics disc), special care such as ICU stay, admission service (admit service), transfusion requirement (transfusions), secondary insurer (insurance 2), and grouped age. The discriminative power of these variables was evaluated through Weight of Evidence (WoE) distributions, and the optimized categories are displayed in the supplementary material figure SX.

\subsection{Model Performance}

The model's performance was evaluated using multiple metrics for both the test and validation datasets shown in Table \ref{tab:performance}

\begin{table}
\small
\begin{tabular}{|c|c|c|}%
\hline%
Measure&Test set (CI 95\%)&Validation set (CI 95\%)\\%
\hline%
Accuracy&0.71 (0.71, 0.72)&0.76 (0.76, 0.77)\\%
Positive Predictive value or Precision&0.61 (0.6, 0.62)&0.67 (0.66, 0.69)\\%
Sensitivity or Recall&0.69 (0.68, 0.7)&0.64 (0.62, 0.65)\\%
F1 Score&0.64 (0.64, 0.65)&0.65 (0.64, 0.67)\\%
True Positive Rate&0.69 (0.68, 0.7)&0.64 (0.62, 0.65)\\%
Specificity or True Negative Rate&0.73 (0.72, 0.74)&0.83 (0.82, 0.84)\\%
AUC ROC&0.78 (0.777, 0.788)&0.82 (0.816, 0.831)\\%
\hline%

\end{tabular}
\caption{Prediction performance}
\label{tab:performance}
\end{table}

\begin{figure}
    \centering
    \includegraphics[width=0.75\textwidth]{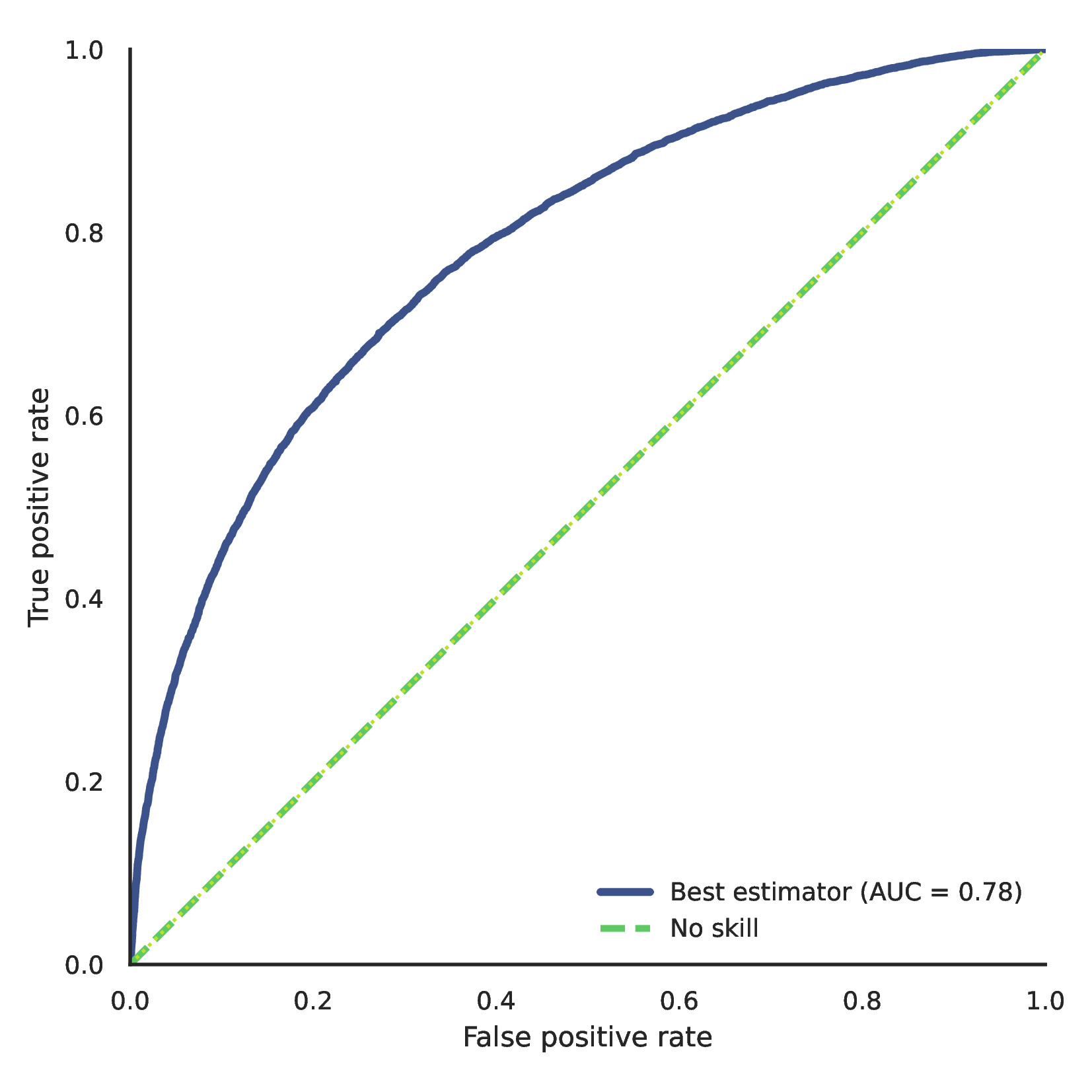}
    \caption{Receiver Operating Characteristic (ROC) curve for the test dataset, illustrating the model’s discriminative performance between short and prolonged hospital stays across different classification thresholds.}
    \label{fig:roc}
\end{figure}

\begin{figure}
    \centering
    \includegraphics[width=0.75\textwidth]{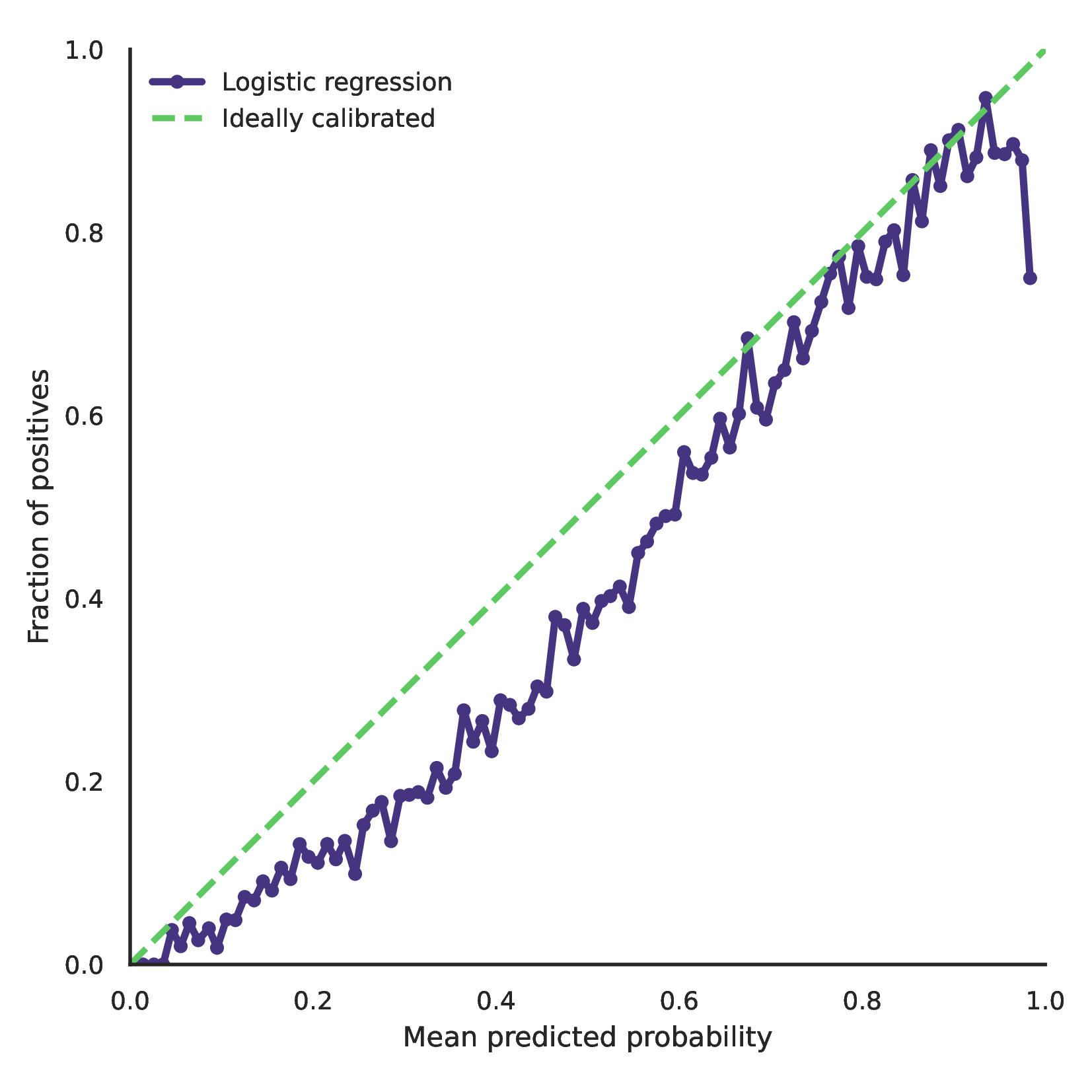}
    \caption{Calibration curve for the test dataset, showing the agreement between predicted probabilities and observed outcomes, indicating a good calibration of the model.}
    \label{fig:calibration}
\end{figure}

\begin{figure}
    \centering
    \includegraphics[width=0.75\textwidth]{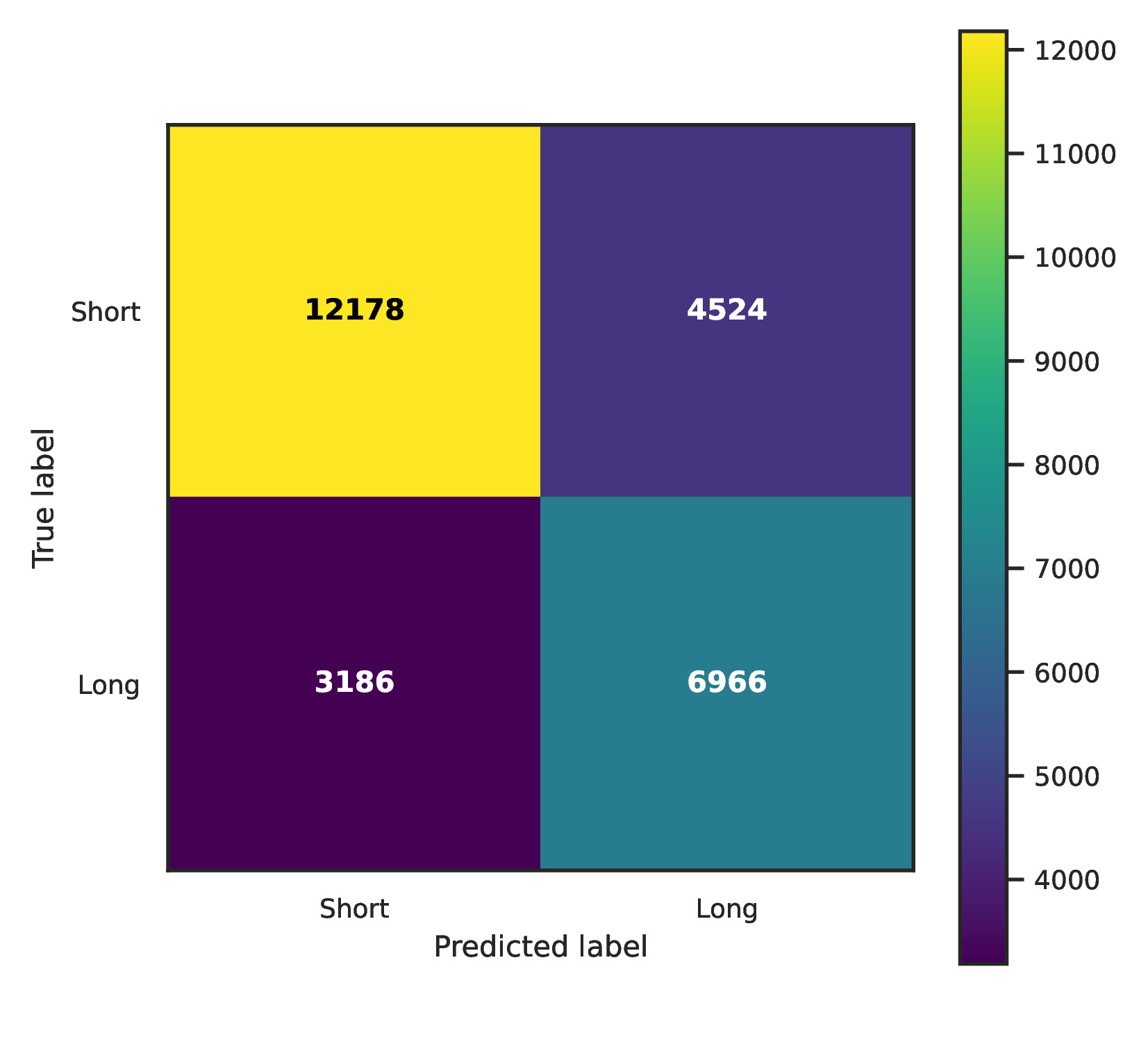}
    \caption{Confusion matrix for the test dataset, summarizing the classification performance of the model for short and prolonged hospital stays.}
    \label{fig:confusion}
\end{figure}

The ROC curve for the test set figure \ref{fig:roc} illustrates the model's ability to distinguish between outcomes. In contrast, the calibration curve figure \ref{fig:calibration} demonstrated a good fit between predicted and observed probabilities. The confusion matrix figure \ref{fig:confusion} showed the model's predictions of short and prolonged stays, with more accurate classification for short stays.

The model's calibration was assessed using the calibration curve shown in figure \ref{fig:calibration}, which presented results closely aligning with the ideal calibration line. The slope and intercept were statistically not significant. This indicated good calibration, with the model's predicted probabilities of prolonged hospital stays matching the observed outcomes. The well-calibrated model suggests that the predictions were reliable across different probability thresholds.

Model performance is competitive with existing studies in the literature like models based on traditional severity scoring systems such as SAPS II (AUC-ROC: 0.70, 0.667) \cite{Zoller2014} \cite{Wu2021}, SOFA score (Sensitivity: 0.71) \cite{Houthooft2015} and EuroSCORE (AUC-ROC 0.729) \cite{Meadows2018}, or predictive models applied in specific populations like logistic regression (AUC-ROC: 0.72, 0.745) \cite{Rotar2022} \cite{Wu2021}, support vector machine (Recall: 0.883) \cite{Houthooft2015} (AUC-ROC: 0.716) \cite{Wu2021} and Artificial neural networks (Recall: 0.775) \cite{Houthooft2015} (AUC-ROC: 0.743) \cite{Wu2021}. A similar in purpose model, all hospitalized patients with various diseases on their first day of admission, have a lower performance (Accuracy: 0.68). 

\subsection{Comparison with standard feature selection approaches}

In comparison with existing approaches, our model introduces a key differentiating element: the integration of feature Information Value and correlation cliques for feature selection. In our setting, when all the admission-level features were included with only the basic selection step, the model have a performance accuracy of 0.75 and a sensitivity or recall of 0.43. Including all the features does not substantially improve beyond the accuracy 0.71 and rather have a lower recall performance compared to the obtained 0.69 with the proposed selection method. The use of such selection improves interpretability and definitely without penalizing predictive model performance.

When compared to more conventional selection procedures, such as recursive feature elimination (RFE), our method further demonstrates its novelty. RFE and other standard approaches often prioritize predictive performance by iteratively removing variables that have a limited contribution, but without taking into consideration their relation to the target. RFE produced model with also 9 features but with quite lower sensitivity or recall (0.42 versus 0.64) and similar accuracy (0.74 vs 0.71) suggesting that simply optimizing variable subsets through recursive procedures does not adequately address redundancy among highly correlated variables.

\subsection{Model Explainability}

\begin{figure}
    \centering
    \includegraphics[width=0.75\textwidth]{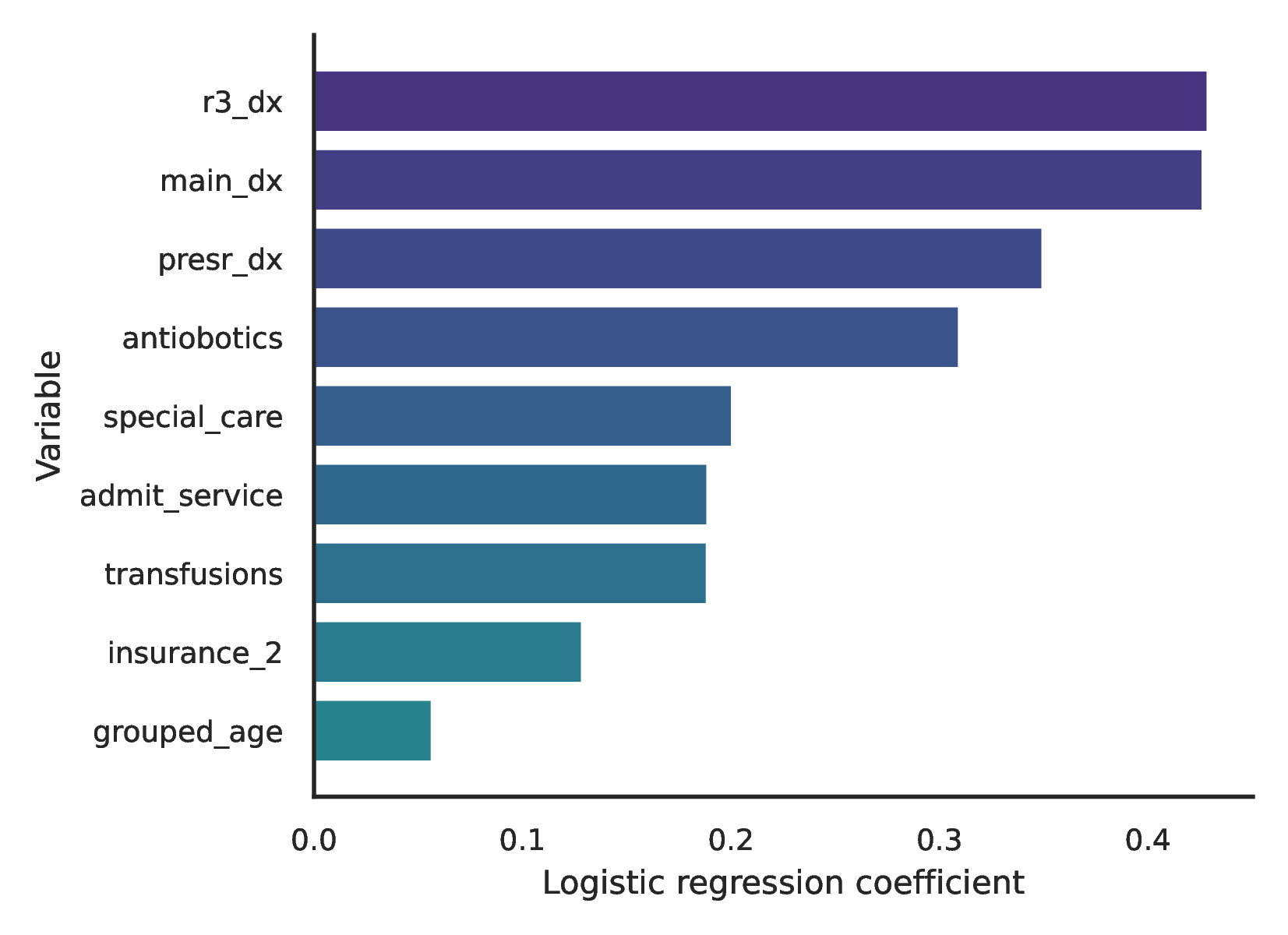}
    \caption{Histogram of the logistic regression coefficients corresponding to the most important features selected by the model. The figure displays the magnitude and direction of the estimated weights, highlighting the reduced set of variables obtained through the feature selection process and supporting the interpretability of the final model.}
    \label{fig:coefficients}
\end{figure}

As mentioned earlier, one of the most important aspects of the model-building process was achieving high interpretability in its results. For this reason, a logistic regression model was chosen, as it offers high interpretability by associating the importance of each variable with the constant assigned in the final model. The weights of our model are presented in figure \ref{fig:coefficients}. Additionally, the significant reduction in the number of variables obtained through the feature selection process enables a model with a low number of variables and assigned importance values, thereby ensuring high interpretability.

Moreover, the use of WoE to assign numerical values to each category/range of values allows for a detailed view of which categories the model associates with long or short stays. In this case, categories with high WoE values correspond to a higher probability of prolonged stays, while negative WoE values correspond to a higher probability of short stays.

\begin{figure}
    \centering
    \includegraphics[width=\textwidth]{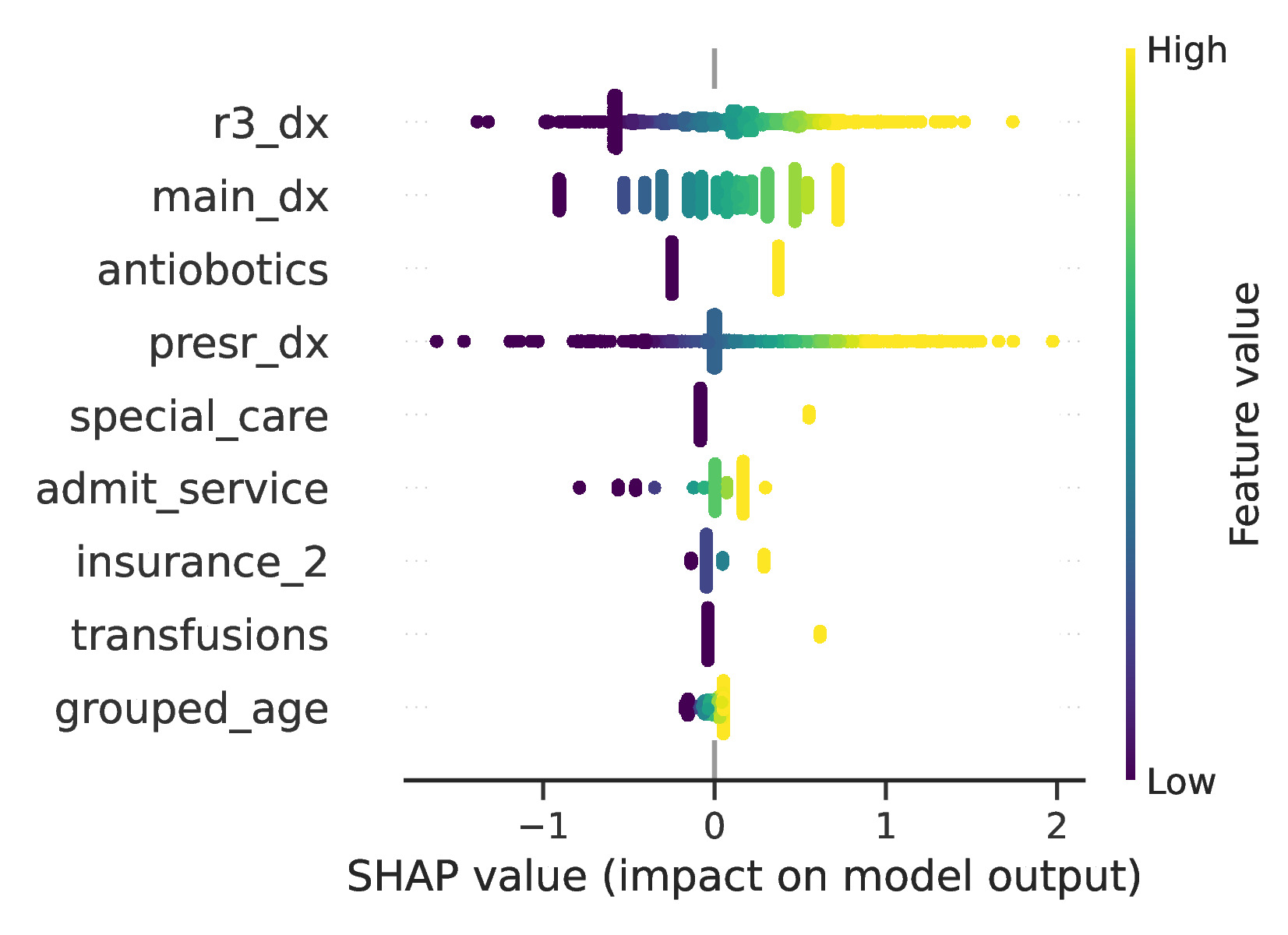}
    \caption{SHAP beeswarm plot for the final model, showing the distribution and magnitude of SHAP values for each feature across all observations. The figure summarizes the global importance and direction of each variable’s contribution to the model predictions, and confirms the consistency of the interpretability results obtained using SHAP with those derived from the WoE-based coefficient analysis.}
    \label{fig:shap}
\end{figure}

In addition to this, a Shapley Additive Explanations (SHAP) \cite{lundberg2020explainable} analysis was performed on the final model to confirm that the interpretability obtained with both methods, WoE and SHAP, is consistent with each other. The results of this analysis are presented in figures \ref{fig:coefficients} and \ref{fig:shap}, where it can be observed that both interpretability values are consistent for all variables in the model.

For the final set of variables, we can see that the first three most important variables correspond to diagnostic variables (r3 dx, main dx, presr dx). The next six variables are treatment variables (antibiotics, transfusions, special care, admit service) with a completely external variable (insurance 2) and only one demographic variable having less weight (grouped age). This small group of variables, including the most important diagnostic variables, has enough information to discriminate pLoS.

\subsection{Discussion} 

The model was validated using data obtained outside the period of time defined for the training and testing datasets. The calibration curve demonstrated a good fit compared to the ideal calibration, indicating the model's reliable performance. The datasets’ features were reduced from 52 to 9, increasing the model’s interpretability. The main diagnosis (main dx) and related R3 diagnosis (r3 dx) features were the most relevant for predicting prolonged lengths of stay. The Information Value (IV), used to describe these features, proved to be an effective method when features were associated with continuous variables. However, for categorical features with a large number of categories (such as the main diagnosis), optimal binning provided a more accurate description. The use of Weight of Evidence (WoE) enabled a detailed view of how specific categories are linked to prolonged or short stays.  The results from both the WoE and SHAP analyses were consistent with each other, indicating that the model's interpretability was robust across all variables. Additionally, the importance of each feature was also visually confirmed through SHAP values, making the model highly interpretable for clinical application.
The model demonstrated strong performance with an AUC ROC of $0.82$ ($95\% CI: 0.816-0.831$) on the validation set. This level of performance was achieved with only 9 variables, distinguishing our approach from other models that often rely on more specific and larger feature sets for specialized hospital populations. This model was applied to a broader population, offering a more generalized solution that maintains high efficacy while simplifying the feature set. This reduction not only enhances the model’s performance by focusing on variables with the most information, but also reduces training costs. The ability to identify key variables and their importance, along with numerical values, facilitates an understanding of the factors influencing predictions, which is essential in the medical field. Additionally, the versatility of the feature selection method allows for its application to different hospital datasets, including both categorical and numerical data, ensuring the identification of key variables relevant to various populations.

\subsection{Limitations and usability}

The information was obtained from a database curated at the end of the patient's stay. This means that the information used to build the model is of high quality. This quality will not be present in possible future real-time studies; therefore, special treatment in such a setting will need to be devised. An example of such methods is the algorithms to handle missing data in real time LoS predictions \cite{Pauphilet2020}.

We want to note that the interpretation derived from WOE and IV quantities is not at the level of a causality model, and such use should be avoided. We think that a procedure where WOE values are assigned to feature categorical values corresponds to a transfer of Los information (target) to the variable, which helps to increase prediction performance. To determine the most effective intervention tools, additional causal inference methods should be employed.

The model will be utilized by hospital staff in future prospective research to validate its performance and investigate whether it functions as expected in real-world settings, thereby mitigating risks associated with unforeseen challenges during deployment. No patient will interact with the model or the data.

\section{Conclusions}

The model demonstrates  strong performance, achieving comparable discrimination metrics to those of leading studies in the field, with an AUC-ROC of 0.82 (0.816-0.831) on the validation set. In particular, this level of performance is achieved with a distinctly reduced number of variables, distinguishing our approach from many similar studies that often focus on more specific hospital populations. By applying our model to a broader population, we offer a more generalized solution that maintains high efficacy while significantly simplifying the feature set. This underscores the model's robustness and its potential for broader applicability in real-world settings.

We introduced an innovative feature selection algorithm that combines WOE, IV, and clique algorithms, resulting in a significant reduction in the number of variables while keeping only the most relevant ones. This reduction not only enhances model performance and training costs by focusing on the most variables with more information but also improves the interpretability of the results. The ability to identify key variables with assigned importance values facilitates a clearer understanding of the factors influencing predictions, which is crucial for applications in the medical field. In future studies or deployments, this interpretability will contribute to better integration into clinical reasoning, providing a robust tool for informed decision-making.

The versatility of our feature selection method enables its application to various hospital datasets, facilitating the identification of key variables for different populations. This capability ensures that the models developed are both relevant and interpretable for each specific context. Applying this method to various populations not only improves the accuracy of predictions but also facilitates more targeted resource allocation, potentially leading to reductions in hospital length of stay and enhanced management of patient care.

\section{Acknowledgement}
The authors acknowledge funding from Proyecto Innovación Abierta Hospital Alma Máter de Antioquia y Universidad de
Antioquia

\newpage

\section{References}
\bibliographystyle{elsarticle-num} 
\bibliography{refs}

\end{document}